\documentclass[sigconf]{acmart}

\usepackage{pifont}
\usepackage{multirow}
\usepackage[table,xcdraw]{xcolor}
\usepackage{tikz}
\usepackage{bm}

\AtBeginDocument{%
  }

\copyrightyear{2025}
\acmYear{2025}
\setcopyright{acmlicensed}
\acmConference[MM '25] {Proceedings of the 33rd ACM International Conference on Multimedia}{October 27--31, 2025}{Dublin, Ireland.}
\acmBooktitle{Proceedings of the 33rd ACM International Conference on Multimedia (MM '25), October 27--31, 2025, Dublin, Ireland}
\acmISBN{979-8-4007-2035-2/2025/10}
\acmDOI{10.1145/3746027.3755413}

\acmSubmissionID{3644}

\begin{document}

\title{DINOv2 Driven Gait Representation Learning for Video-Based Visible-Infrared Person Re-identification}

\author{Yujie Yang}
\authornote{Equal Contribution.}
\orcid{0009-0006-4866-1117}
\affiliation{%
  \institution{Kunming University of Science and Technology}
  \department{Faculty of Information Engineering and Automation}
  \city{Kunming}
  \state{Yunnan}
  \country{China}
}
\email{20232104053@stu.kust.edu.cn}

\author{Shuang Li}
\authornotemark[1]
\orcid{0000-0003-4258-3163}
\affiliation{%
  \institution{Chongqing University of Post and Telecommunications}
  \department{School of Computer Science and Technology}
  \city{Chongqing}
  \state{Chongqing}
  \country{China}
}
\email{shuangli936@gmail.com}

\author{Jun Ye}
\orcid{0009-0001-6754-8578}
\affiliation{%
  \institution{China University of Mining Technology}
  \department{School of Information and Control Engineering}
  \city{Xuzhou}
  \state{Jiangsu}
  \country{China}
}
\email{tb22060028a41@cumt.edu.cn}

\author{Neng Dong}
\orcid{0000-0001-5523-1082}
\affiliation{%
  \institution{Nanjing University of Science and Technology}
  \department{School of Computer Science and Engineering}
  \city{Nanjing}
  \state{Jiangsu}
  \country{China}
}
\email{neng.dong@njust.edu.cn}

\author{Fan Li}
\authornote{Corresponding author.}
\orcid{0000-0002-3883-0275}
\affiliation{%
  \institution{Kunming University of Science and Technology}
  \department{Faculty of Information Engineering and Automation}
  \city{Kunming}
  \state{Yunnan}
  \country{China}
}
\email{20150032@kust.edu.cn}

\author{Huafeng Li}
\orcid{0000-0003-2462-6174}
\affiliation{%
  \institution{Kunming University of Science and Technology}
  \department{Faculty of Information Engineering and Automation}
  \city{Kunming}
  \state{Yunnan}
  \country{China}
}
\email{hfchina99@163.com}

\renewcommand{\shortauthors}{Yujie Yang et al.}

\begin{abstract}
Video-based Visible-Infrared person re-identification (VVI-ReID) aims to retrieve the same pedestrian across visible and infrared modalities from video sequences. 
Existing methods tend to exploit modality-invariant visual features but largely overlook gait features, which are not only modality-invariant but also rich in temporal dynamics, thus limiting their ability to model the spatiotemporal consistency essential for cross-modal video matching. 
To address these challenges, we propose a DINOv2-Driven Gait Representation Learning (DinoGRL) framework that leverages the rich visual priors of DINOv2 to learn gait features complementary to appearance cues, facilitating robust sequence-level representations for cross-modal retrieval. 
Specifically, we introduce a Semantic-Aware Silhouette and Gait Learning (SASGL) model, which generates and enhances silhouette representations with general-purpose semantic priors from DINOv2 and jointly optimizes them with the ReID objective to achieve semantically enriched and task-adaptive gait feature learning. 
Furthermore, we develop a Progressive Bidirectional Multi-Granularity Enhancement (PBMGE) module, which progressively refines feature representations by enabling bidirectional interactions between gait and appearance streams across multiple spatial granularities, fully leveraging their complementarity to enhance global representations with rich local details and produce highly discriminative features. 
Extensive experiments on HITSZ-VCM and BUPT datasets demonstrate the superiority of our approach, significantly outperforming existing state-of-the-art methods.

\end{abstract}






\maketitle
\begin{figure}[t!]
\centering
\includegraphics[width=8.8cm,keepaspectratio=true,page=1]{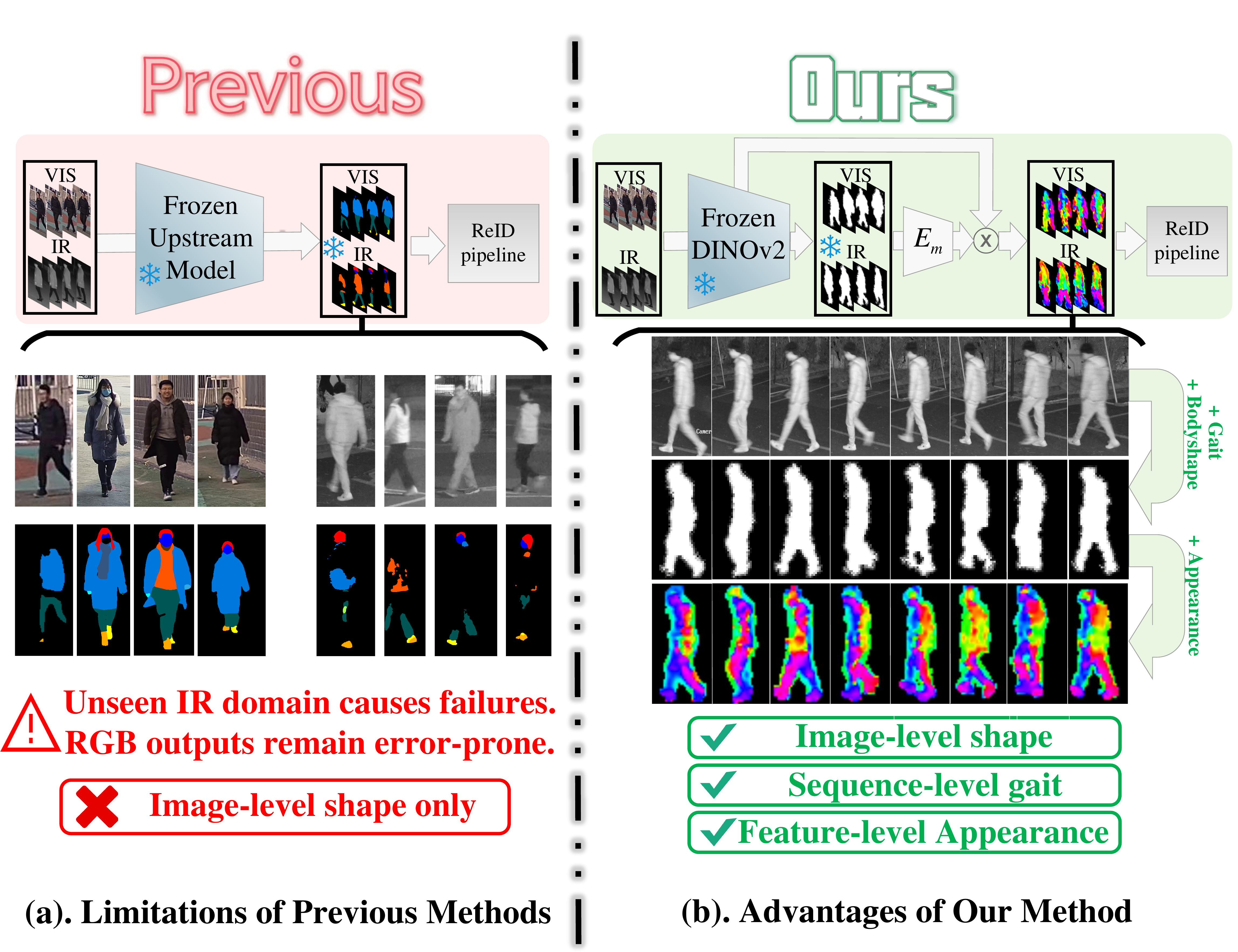}
\caption{
Motivation of DinoGRL. (a) Existing shape-based VI-ReID methods often rely on image-level parsing networks, which are not optimized for ReID, particularly under the infrared modality——leading to noisy segmentation and neglect of temporal gait cues. (b) In contrast, DinoGRL leverages DINOv2 as a strong visual prior to produce high-quality silhouettes that facilitate the integration of sequence-level gait features, further refined by complementary appearance cues to achieve discriminative and modality-robust embeddings.
}
\label{motivation1}
\end{figure}

\section{Introduction}

Visible-Infrared person re-identification (VI-ReID) \cite{VIreid1, VIreid2, VIreid3, VIreid4, VIreid5} has garnered increasing attention due to its ability to match person images across different modalities, enabling robust identification under varying illumination conditions for around-the-clock surveillance. However, existing VI-ReID methods primarily focus on static image matching, which limits their ability to leverage spatiotemporal consistency and fine-grained motion cues inherent in real-world scenarios. To address these limitations, Video-based Visible-Infrared person ReID (VVI-ReID) \cite{VVIreid1, IBAN, VVIreid3, VVIreid4, VCM, BUPT, li2025video} has recently emerged as a promising direction. By incorporating temporal dynamics and cross-modal alignment, VVI-ReID enhances retrieval performance in complex surveillance environments.

VVI-ReID aims to learn modality-invariant and temporally consistent representations for accurate pedestrian matching across visible(VIS) and infrared(IR) video sequences. Existing approaches typically align features from different modalities within a shared embedding space to learn modality-invariant representations\cite{VIshare1, VIshare2, VIshare3, li2023logical}, and attempt to mitigate modality discrepancies by incorporating auxiliary information (e.g., shape) to provide additional guidance for cross-modal feature alignment\cite{VIsp1, VIsp2, IBAN, li2023shape, wang2022body, zhang2022cross}. 
Despite the remarkable progress of existing methods, they often overlook discriminative sequence-level gait patterns that encode crucial temporal dynamics. Gait is represented as a sequence of body silhouettes that inherently capture temporal dynamics and exhibit strong robustness to modality variations. It reflects the unique walking pattern of an individual and has shown remarkable effectiveness in retrieval tasks \cite{fan2023opengait, ye2024biggait, ye2025biggergait_neurips, leng2025dual, dong2023erasing, CFine, MANet, LCR2S}. While gait offers robust and temporally rich cues, existing VVI-ReID methods generally overlook such information. Moreover, shape-based VI-ReID methods \cite{li2023shape, hong2021fine, huang2022cross} cannot be directly applied in the VVI-ReID task, as they still face significant limitations, as shown in Fig. 1 (a): 
\textbf{(1) Neglect of Sequence-Level Gait Patterns. }Exist methods focus exclusively on image-level shape while neglect discriminative sequence-level gait patterns, thereby limiting their capacity to model temporal dynamics critical for video tasks. \textbf{(2) Neglect of Appearance information deficiency in Silhouette Representations. }Silhouette maps lack detailed appearance textures, which are complementary to gait and important for fine-grained identity matching. \textbf{(3) Dependence on Non-ReID-Optimized Upstream Models. }Current methods rely heavily on upstream models that are not optimized for ReID, particularly in the IR modality. This leads to poor-quality silhouette maps and, in turn, degrades the overall performance of downstream ReID tasks.

To address these limitations and integrate gait into VVI-ReID, a more powerful and generalizable visual representation is needed, it should not only capture temporal gait and shape semantics but also preserve fine-grained texture details. 
Inspired by BigGait\cite{ye2024biggait} and BiggerGait\cite{ye2025biggergait_neurips}, which together demonstrate the strong potential of large vision models in learning robust and discriminative gait representations, we further extend this idea to the VVI-ReID. Specifically, we introduce \textbf{DINOv2}\cite{oquab2023dinov2}, a large-scale vision model pretrained on web-scale data without task-specific supervision, to provide a more general and transferable visual foundation.
Owing to its exposure to diverse visual tasks, including classification, segmentation, depth estimation, and retrieval, DINOv2 learns rich and generalizable visual representations that capture both global structure and fine-grained local details.
Such capabilities make it well-suited for generating high-quality gait representations, even from noisy or low-resolution IR silhouettes, effectively alleviating the silhouette degradation issue highlighted in Fig. 1(a).
It is worth noting that appearance texture and shape \& gait are inherently complementary. Our objective is not only to obtain high-quality gait representations, but also to fully exploit their complementarity through targeted mutual enhancement. Their synergistic interaction enables more robust represent, significantly boosting VVI-ReID performance.

Based on the motivations discussed above, we propose the DINOv2-Driven Gait Representation Learning (DinoGRL) framework, which consists of two branches for learning appearance and gait features, respectively. To facilitate gait feature learning, we introduce the Semantic-Aware Silhouette and Gait Learning (SASGL) module. 

Compared with previous methods, SASGL first leverages DINOv2's general-purpose visual priors to generate high-quality, semantically enriched silhouette maps for robust gait feature extraction. Furthermore, by jointly optimizing these representations with the ReID objective, SASGL enables the learning of gait features that are not only modality-invariant but also task-adaptive.
Specifically, SASGL first extracts an initial pedestrian mask using semantic features from DINOv2’s final layer, and then enriches it by incorporating intermediate features that encode multi-level semantic information, enabled by the general-purpose visual priors learned through DINOv2’s diverse task pretraining, to effectively compensate for the loss of appearance textures.
To fully exploit the complementary strengths of appearance and gait features, we propose the Progressive Bidirectional Multi-Granularity Enhancement (PBMGE).  
PBMGE progressively enhances global representations by leveraging fine-grained local interactions across multiple spatial granularities, gradually integrating information from different levels to refine holistic identity cues. This progressive enhancement mechanism enables more precise and robust feature representations compared to conventional direct fusion strategies.

Here are the main contributions of our paper:
(1) We pioneer a synergistic framework for VVI-ReID that extracts general-purpose priors from the task-agnostic DINOv2 and leverages the complementary strengths of appearance and gait to achieve discriminative and modality-robust representations. 
(2) 
SASGL, a silhouette and gait learning module built upon a large vision model, utilizes DINOv2’s general-purpose semantic features to produce high-quality gait representations, which are adaptive to downstream ReID tasks. 
(3) We design the PBMGE module, which enables fine-grained local-to-global compensation and progressive bidirectional enhancement, fully exploiting the complementary characteristics of appearance and gait. 
(4) Extensive experiments on the HITSZ-VCM and BUPT datasets demonstrate that DinoGRL framework achieves new state-of-the-art performance, validating its effectiveness.

\section{RELATED WORK}

\begin{figure*}[t!]
\centering
\includegraphics[width=0.95\textwidth,keepaspectratio=true]{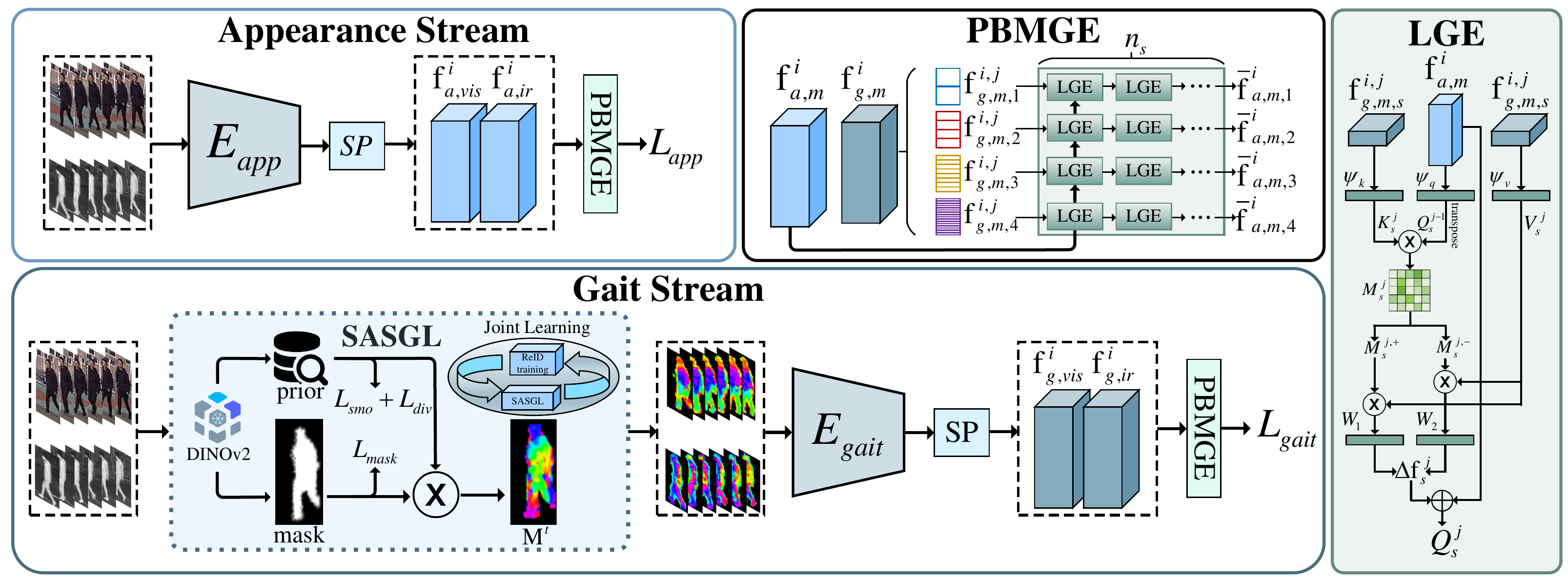}
\caption{The overall framework of DinoGRL. This framework consists of two key modules: SASGL and PBMGE. SASGL employs a Semantic-Aware Silhouette Generator to produce modality-invariant silhouettes, leveraging the general-purpose visual priors of DINOv2 to facilitate gait representation learning. A Joint Learning Strategy is applied to simultaneously optimize silhouette generation and gait feature extraction, yielding gait representation $\mathbf{M}^t$.
PBMGE further enhances global appearance and gait representations by integrating local features from the complementary stream across multiple granularities, yielding robust and discriminative pedestrian embeddings.
}
\label{framework1}
\end{figure*}

\subsection{Video-based Visible-Infrared Person Re-Identification}
Video-based Visible-Infrared Person Re-Identification (VVI-ReID) face two key challenges: bridging the modality gap between RGB and infrared images, and effectively exploiting temporal information.
To address the modality discrepancy, VVI-ReID methods often draw upon advances in visible-infrared person ReID (VI-ReID), which mainly follow two paradigms. The first focuses on learning modality-invariant features through architectural designs \cite{VIshare1, VIshare2, VIshare3}. For example, HSME \cite{hao2019hsme} separates domain-specific and domain-shared layers, while MCSL exploits relationships across cross-modality pairs. 
The second paradigm incorporates auxiliary modality cues to ease cross-modal alignment. HOS-Net \cite{qiu2024high} aligns intermediate features, and Li et al. \cite{li2020infrared} introduce an auxiliary X modality to enhance representation learning.

Building upon these approaches, VVI-ReID further emphasizes temporal modeling for robust video-based matching. Specifically, IBAN \cite{IBAN} utilizes anaglyph images as an auxiliary modality to bridge the modality gap and employs an LSTM to capture temporal dependencies. SAADG \cite{zhou2023video} applies adversarial strategy-based data augmentation to improve sequence-level representations. CST \cite{feng2024cross} adopts a ViT-based architecture to model global spatial-temporal features and long-range dependencies across frames.

\subsection{Upstream Anatomical Modeling for ReID}
Pedestrian walking videos often suffer from background clutter and foreground variations, motivating the use of task-specific representations such as binary silhouettes \cite{hong2021fine}, body skeletons \cite{rao2023transg}, and human parsing maps \cite{he2020grapy, li2020self}. Early works like SPReID \cite{kalayeh2018human} and EaNet \cite{huang2018eanet} leveraged human parsing to suppress background noise and improve feature localization. P2Net \cite{guo2019beyond} and ISP \cite{zhu2020identity} further modeled human body parts and contextual information to enhance discriminability. Recent methods, such as SEFL \cite{feng2023shape} and SCRL \cite{li2023shape}, focus on shape-based feature disentanglement and augmentation. however, they still struggle to fundamentally address the intrinsic shape information degradation under the infrared modality.

\subsection{DINOv2: Self-Supervised Learning of General-Purpose Visual Features}

DINOv2 \cite{oquab2023dinov2} represents a significant advancement in self-supervised learning, demonstrating that large-scale pretraining on curated data can produce universal visual features that generalize across diverse tasks without task-specific fine-tuning. Recent works have leveraged DINOv2 to enhance downstream vision tasks across a variety of downstream vision tasks: 
ViT-CoMer \cite{xia2024vit} integrates convolutional modules into DINOv2 backbones to improve local and multi-scale representations for detection and segmentation; Virchow \cite{vorontsov2024foundation} and Prov-GigaPath \cite{xu2024whole} validate DINOv2’s generalization to medical imaging tasks; RoMa \cite{edstedt2024roma} and SALAD \cite{izquierdo2024optimal} adapt DINOv2 features for dense matching and visual place recognition, respectively. 
BigGait\cite{ye2024biggait} and BiggerGait\cite{ye2025biggergait_neurips} validates the potential of DINOv2 for learning robust and discriminative gait representations.
These successes highlight DINOv2’s strong potential as a task-agnostic feature extractor, motivating its adoption in our framework for robust gait representation learning.

\section{METHODOLOGY}
In this section, we present the implementation details of our DinoGRL framework, with an overview illustrated in Fig.~\ref{framework1}.

\subsection{Appearance Representation Learning}\label{sec3.1}
Given the sample set $D = \{X_{vis}^i, X_{ir}^i\}_{i=1}^{N}$, where $X_m^i = \{X_m^{i,t} \mid X_m^{i,t} \in \mathbb{R}^{C \times H \times W}\}_{t=1}^{T}$ denotes the $i$-th input sequence from modality $m \in {\text{vis}, \text{ir}}$, and $C$, $H$, and $W$ represent the number of channels, height, and width, respectively, while $T$ is the number of frames. Each sequence $\{X_m^{i,t}\}_{t=1}^{T}$ is first processed by the appearance encoder $\bm{E}_{app}$, following the AGW design~\cite{ye2021deep,VCM}, and subsequently aggregated using Set Pooling (SP)~\cite{chao2019gaitset} to produce the sequence-level feature representation:
\begin{equation} 
\textbf{f}\hspace{0.1em}_{a,m}^i = SP(\bm{E}_{app}(\{X_m^{i,t}\}_{t=1}^{T})), \end{equation} 
where $i$ indexes the sample within a mini-batch.
To ensure the identity discriminability of $\textbf{f}\hspace{0.1em}_{a,m}^i$, we adopt the Cross-Entropy loss $\mathcal{L}_{id}^{app}$ and the triplet loss $\mathcal{L}_{tri}^{app}$ as supervision, formulated as: 
\begin{equation} \mathcal{L}_{id}^{app} = -q_i \log(\bm{W}_{id}(\textbf{f}\hspace{0.1em}_{a,m}^{i})), \end{equation} 

\begin{equation}
\mathcal{L}^{tri}_{app} = \left[ c + D_{a,p} - D_{a,n} \right]_+,
\end{equation}
where $\bm W_{id}$ denotes the shared identity classifier for IR and VIS features, $q_{i}\in\mathbb{R}^{K \times 1}$ is a one-hot vector, and only the element at $y_{i}$ is $1$. For $\mathcal{L}_{tri}^{app}$, $D_{a,p}$ and $D_{a,n}$ denote the squared Euclidean distances between the anchor $a$ and the positive sample $p$, and between the anchor $a$ and the negative sample $n$, respectively. $c$ is a margin that enforces a minimum distance between positive and negative pairs. $a$, $p$, and $n$ represent the indices of the anchor, positive, and negative samples, respectively. $[z]_+$ equals to $max(z,0)$.

However, as discussed in the introduction, directly extracting sequence-level appearance features from pedestrian sequences is susceptible to modality variations. This leads to unstable representations that fail to capture consistent biometric cues, such as gait patterns, which are inherently modality-invariant. As a result, the model's performance is fundamentally limited. To address this issue, we further introduce a \textbf{Semantic-Aware Silhouette and Gait Learning} module, aiming to enhance the cross-modality robustness by explicitly modeling gait-related semantic features.

\subsection{Semantic-Aware Silhouette and Gait Learning}\label{sec3.2}
To explicitly model gait features, a straightforward approach is to utilize human semantic parsing models (e.g., SCHP \cite{li2020self}, Grapy-ML \cite{he2020grapy}) to parse pedestrian images in sequential frames, thereby generating continuous silhouette maps for subsequent gait feature extraction. However, this strategy faces two major limitations. First, conventional human semantic parsing models exhibit insufficient visual representation capabilities, leading to noticeable defects in the generated silhouettes, particularly under the infrared modality where parsing quality degrades significantly (as illustrated in Fig. \ref{motivation1}). Second, these parsing models are not originally designed for the person re-identification (ReID) task, and thus the extracted silhouette features are not optimally aligned with ReID objectives. To address these limitations, 
%
inspired by BigGait\cite{ye2024biggait} and BiggerGait\cite{ye2025biggergait_neurips}, a Semantic-Aware Silhouette and Gait Learning (SASGL) module is introduced. 
This module comprises two core components: (1) \textbf{Semantic-Aware Silhouette Generator (SASG)}, which leverages the strong general-purpose visual priors of DINOv2 to enhance the semantic richness and fidelity of silhouette representations; and (2) \textbf{Joint Learning Strategy}, which performs end-to-end optimization with the ReID objective, enabling the silhouette representations to be adaptively aligned with downstream recognition requirements. The detailed design and implementation of these two components are presented in the following subsections.

\begin{figure}[t!]
\centering
\includegraphics[width=8cm,keepaspectratio=true,page=1]{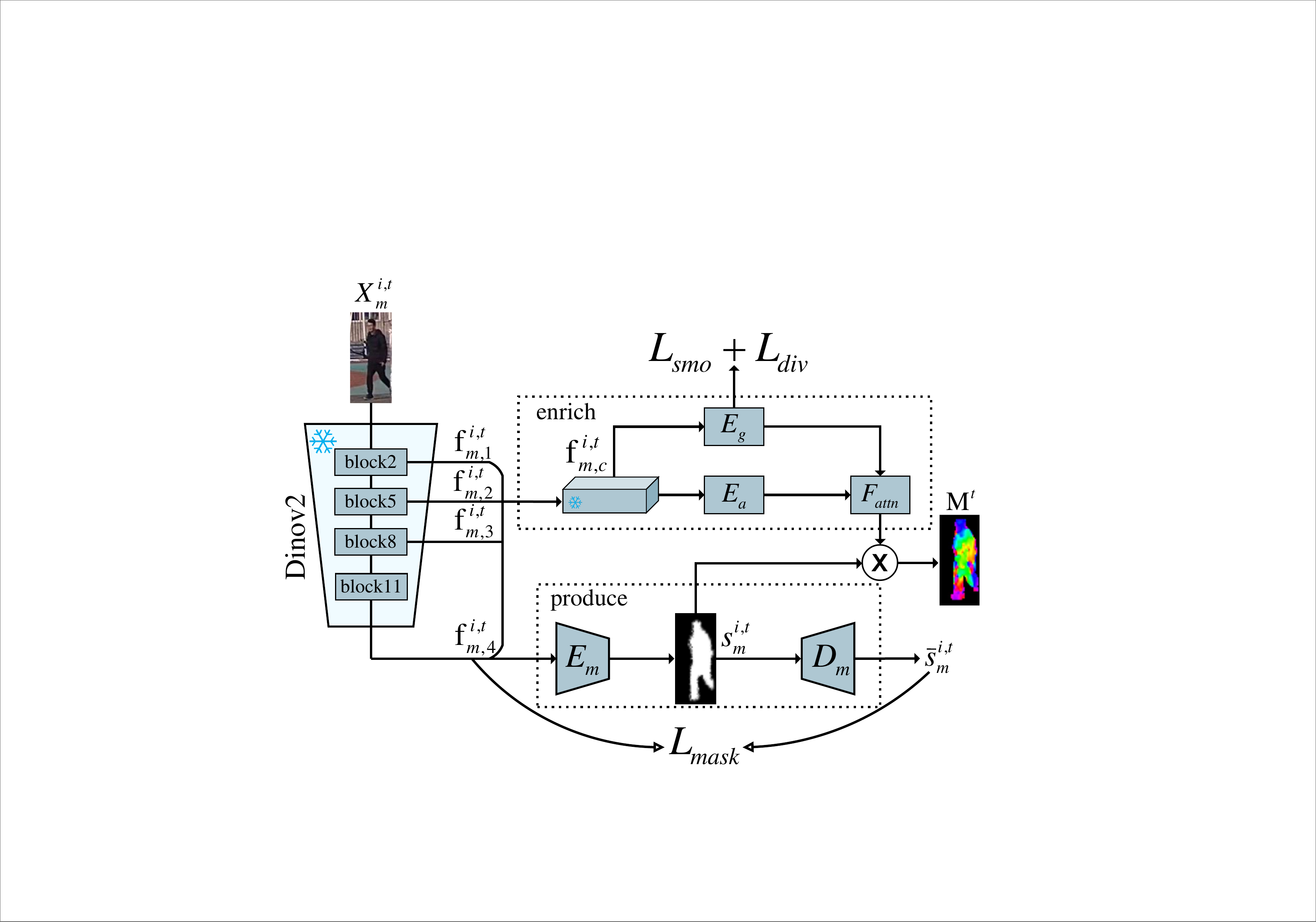}
\caption{Illustration of the 
SASG, which produce and enrich silhouette representations with general-purpose semantic priors from DINOv2. }
\label{SASG}
\end{figure}

\textbf{Semantic-Aware Silhouette Generator (SASG)} aims to generate modality-invariant and semantically enriched silhouette representations. As shown in Fig. \ref{SASG}, SASG is designed with two objectives: (1) producing coherent silhouette masks that preserve stable gait patterns across modalities, and (2) enriching these masks with semantic information from DINOv2’s general-purpose visual priors.

For the first objective, we feed the input sequence $\{X_m^{i,t}\}_{t=1}^{T}$ into the DINOv2 backbone and utilize the highest-level semantic feature maps $\{\mathbf{f}_{m,4}^{i,t}\}_{t=1}^{T}$ extracted from the final block. Each feature map $\mathbf{f}_{m,4}^{i,t}$ undergoes batch normalization and is projected into 2 channels by an encoder $\bm E_m$, implemented as a linear convolutional layer. A softmax operation partitions each $\mathbf{f}_{m,4}^{i,t}$ into foreground and background components, and the spatially centered foreground mask $S^t_{m}$ is selected as the pedestrian silhouette for each frame:
\begin{equation}
S^{i,t}_{m} = \sigma(\bm E_m(\text{BN}(\mathbf{f}_{m,4}^{i,t}))),
\end{equation}
where $\mathbf{f}_{m,4}^{i,t} \in \mathbb{R}^{HW \times C}$ and $\sigma(\cdot)$ denotes the softmax activation applied along the channel dimension.
Since DINOv2 is frozen during training, $\mathbf{f}_{m,4}^{i,t}$ remains static. To preserve the semantic prior of DINOv2 while adapting the mask representations to the downstream ReID task, we introduce a regularization loss $\mathcal{L}_{mask}$, which forces the decoded mask features to remain close to the original $\mathbf{f}_{m,4}^{i,t}$:
\begin{equation}
\mathcal{L}_{mask} = \frac{1}{T} \sum_{t=1}^{T} \| \overline{S}^{i,t}_{m} - \mathbf{f}_{m,4}^{i,t} \|_2,
\end{equation}
where $\overline{S}^{i,t}_{m} = D_m(S^{i,t}_{m})$, and $D_m$ is a linear decoder restoring the channel dimension to 384.

For the second objective, we extract multi-level feature maps $\{\mathbf{f}_{m,1}^{i,t}, \mathbf{f}_{m,2}^{i,t}, \mathbf{f}_{m,3}^{i,t}, \mathbf{f}_{m,4}^{i,t}\}_{t=1}^{T}$ from the 2nd, 5th, 8th, and final blocks of DINOv2, respectively, corresponding to progressively increasing semantic levels. These multi-level features are concatenated along the channel dimension for each frame to form unified representations $\{\textbf{f}_{m,c}^{i,t}\}_{t=1}^{T}$, where $\mathbf f_{m,c}^{i,t} \in \mathbb{R}^{HW \times 4C}$.
Subsequently, two parallel encoders, $\bm E_g$ and $\bm E_a$, are employed to extract gait-specific and appearance-specific priors from each $\mathbf{f}_{m,c}^{i,t}$, respectively. Their outputs are then fused through an attention mechanism to produce enriched silhouette features:
\begin{equation}
\mathbf{M}^t = S^t_{m} \times \mathcal{\bm F}_{attn}(\bm E_g(\mathbf{f}_{m,c}^{i,t}), \bm E_a(\mathbf{f}_{m,c}^{i,t})),
\end{equation}
where $\mathcal{\bm F}_{attn}(\cdot, \cdot)$ denotes the attention-based fusion module\cite{fan2024skeletongait}.
To encourage $\bm E_g$ and $\bm E_a$ to specialize in gait and appearance prior extraction, we introduce two regularization terms.
First, a smoothness loss $\mathcal{L}_{smo}$ promotes spatial consistency by penalizing spatial gradients of $\bm E_g(f_{m,c}^{i,t})$:
\begin{equation}
\mathcal{L}_{smo} = \frac{1}{T} \sum_{t=1}^{T} \left( \left| \text{sobel}_x \times \bm E_g(\mathbf{f}_{m,c}^{i,t}) \right| + \left| \text{sobel}_y \times \bm E_g(\mathbf{f}_{m,c}^{i,t}) \right| \right),
\end{equation}
where $\text{sobel}_x$ and $\text{sobel}_y$ are Sobel operators\cite{sobel19683x3} along the x- and y-axes.
Second, a diversity loss $\mathcal{L}_{div}$ prevents feature collapse by maximizing the entropy of channel activation distributions: 
\begin{equation}
\mathcal{L}_{div} = \frac{1}{T} \sum_{t=1}^{T} (H_{\text{max}} - H(P^{i,t}_{m})),
\end{equation}
where $H(P^{i,t}_m) = -\sum_{i=1}^{C} P_{m,i}^{i,t} \log(P_{m,i}^{i,t})$, $H_{\text{max}}$ is the maximum achievable entropy. The channel activation probability per frame is normalized as: $P_{m,i}^{i,t} = {\sum\limits_{j=1}^{HW} \text{sum}(\bm E_g(f_{m,c}^{i,t,j}))} 
/\
{\sum\limits_{j=1}^{CHW} \text{sum}(\bm E_g(f_{m,c}^{i,t,j}))}$.

\textbf{Joint Learning Strategy}
aims to align the silhouette generation process with the downstream ReID objective by jointly optimizing the Semantic-Aware Silhouette Generator (SASG) and the gait feature extraction network under unified ReID supervision.
Specifically, the generated silhouette  $\{\textbf{M}\hspace{0.1em}^t\}_{t=1}^{T}$ are fed into the gait feature extractor $\bm{E}_{gait}$ 
to produce identity embeddings ${\textbf{f}\hspace{0.1em}^i_{g}}$. Meanwhile, the components of SASG, including $\bm E_m $, $\bm E_g$, and $\bm E_a$, are jointly optimized together with \(\bm{E}_{gait}\) under the supervision of ReID losses similar to Eq (1) and (2), formulated as:
\begin{equation}
\mathcal{L}_{gait} = \mathcal{L}^{id}_{gait} + \mathcal{L}^{tri}_{gait},
\end{equation}
where \(\mathcal{L}_{id}\) denotes the identity loss and \(\mathcal{L}_{tri}\) denotes the triplet loss.
Thanks to the end-to-end optimization design, the gradients of the ReID losses can be backpropagated through \(\bm{E}_{gait}\) to the SASG module via the chain rule. Formally, the gradients with respect to the SASG parameters \(\theta_{SASG}\) are computed as:
\begin{equation}
\frac{\partial \mathcal{L}_{gait}}{\partial \theta_{SASG}} = \frac{\partial \mathcal{L}_{gait}}{\partial \textbf{f}\hspace{0.1em}_g} \times \frac{\partial \textbf{f}\hspace{0.1em}_g}{\partial \theta_{SASG}},
\end{equation}
where \(\theta_{SASG}\) denotes the set of trainable parameters in SASG. This mechanism enables SASG to adaptively refine its outputs during training, ensuring that the generated silhouette representations are progressively aligned with the downstream ReID objective.

\subsection{Progressive Bidirectional Multi-Granularity Enhancement}\label{sec3.3}
While the gait and appearance streams individually capture complementary aspects of pedestrian identity, their independent learning leads to suboptimal feature representations. Specifically, gait features, although modality-invariant, may lack detailed spatial textures, whereas appearance features, though rich in fine-grained details, are vulnerable to modality-induced noise. This motivates the need for effective cross-modal interaction that can leverage the complementary strengths of both streams.
However, directly enhancing features at the local stripe level may lead to fragmented or inconsistent representations across different regions. To address this, we further advocate aggregating the locally enhanced features into global representations at each spatial granularity. 

Thus, we propose Progressive Bidirectional Multi-Granularity Enhancement (PBMGE) module that jointly exploits local and global interactions across multiple spatial granularities. 
Specifically, the inputs to the PBMGE module are the sequence-level features ${\textbf{f}\hspace{0.1em}_{a,m}^i}$ and ${\textbf{f}\hspace{0.1em}_{g,m}^i}$, 
these features are first obtained by applying set pooling (SP)~\cite{chao2019gaitset} over the temporal dimension to frame-level features.
To capture identity cues at different semantic scales, each sequence-level feature is partitioned into 2, 4, 8, and 16 horizontal stripes, producing four granularities. We define the granularity index as $s \in \{1,2,3,4\}$, corresponding to 2, 4, 8, and 16 partitions, respectively. 
For each granularity $s$, the number of partitions is denoted as $n_s$, where $n_1=2$, $n_2=4$, $n_3=8$, and $n_4=16$. At each granularity $s$, the partitioned sub-features are denoted as $\{\textbf{f}\hspace{0.1em}^{i,j}_{a,m,s}\}_{j=1}^{n_s}$ and $\{\textbf{f}\hspace{0.1em}^{i,j}_{g,m,s}\}_{j=1}^{n_s}$, where $j$ indexes the split region.

\begin{table*}[!ht]\small\centering
\caption{Performance comparison with the state-of-the-art Re-ID methods on HITSZ-VCM. `R@1', `R@5' and `R@10' denote Rank-1,Rank-5 and Rank-10, respectively. `-' denotes that no reported result is available.}
\label{Tab1}
\begin{tabular}{m{2.3cm}<{\centering}m{2cm}<{\centering}m{1.5cm}<{\centering}m{1cm}<{\centering}m{0.7cm}<{\centering}m{0.7cm}<{\centering}m{0.7cm}<{\centering}m{0.7cm}<{\centering}m{0.7cm}<{\centering}m{0.7cm}<{\centering}m{0.7cm}<{\centering}m{0.7cm}<{\centering}}
\toprule[0.8pt]


\multirow{2}*{Methods} & \multirow{2}*{Reference} & \multirow{2}*{Type} & \multirow{2}*{Seq\_Len} & \multicolumn{4}{c}{\textit{Infrared to Visible}} & \multicolumn{4}{c}{\textit{Visible to Infrared}}  
\\ \cmidrule(lr){5-8} \cmidrule(lr){9-12} 
                            &  &   &  & {R@1}  & {R@5}  & {R@10}  & mAP    & {R@1}  & {R@5} & {R@10}  & mAP           \\ \toprule[0.8pt]

 Lba\cite{park2021learning}       &ICCV'21 &Image  &6    &46.4 &65.3 &72.2 &30.7 &49.3 &69.3 &75.9 &32.4  \\
 MPANet\cite{wu2021discover}       &CVPR'21 &Image  &6    &46.5 &63.1 &70.5 &35.3 &50.3 &67.3 &73.6 &37.8 \\
 VSD\cite{tian2021farewell}        &CVPR'21 &Image  &6    &54.5 &70.0 &76.3 &41.2 &57.5 &73.7 &79.4 &43.5  \\
 CAJ\cite{ye2021channel}           &ICCV'21 &Image  &6    &56.6 &73.5 &79.5 &41.5 &60.1 &74.6 &79.9 &42.8  \\
 SEFL\cite{feng2023shape}          &CVPR'23 &Image  &6    &67.7 &80.3 &84.7 &52.3 &70.2 &82.2 &86.1 &52.5 \\
 \hline
 MITML\cite{VCM}      &CVPR'22  &Video &6    &63.7 &76.9 &81.7  &45.3 &64.5 &79.0 &83.0 &47.7   \\
 IBAN\cite{IBAN}    &TCSVT'23 &Video &6    &65.0 &78.3 &83.0  &48.8 &69.6 &81.5 &85.4 &51.0   \\
 SADSTRM\cite{li2023adversarial}  &Arxiv'23 &Video &6    &65.3 &77.9 &82.7  &49.5 &67.7 &80.7 &85.1 &51.8   \\
 SAADG\cite{zhou2023video}        &ACM MM'23 &Video &6    &69.2 &80.6 &85.0  &53.8 &73.1 &83.5 &86.9 &56.1   \\
 CST\cite{feng2024cross}          &TMM'24  &Video  &6    &69.4 &81.1 &85.8  &51.2 &72.6 &83.4 &86.7 &53.0   \\
 AuxNet\cite{BUPT}         &TIFS'24 &Video  &6     &51.1 &- &-  &46.0 &54.6 &- &- &48.7   \\
 HD-GI\cite{zhou2025hierarchical}         &INFFUS'25 &Video  &6     &\underline{71.4} &\underline{81.7} &\underline{84.9}  &\underline{57.9} &\underline{74.9} &\underline{84.3} &\underline{87.2} &\underline{60.2}   \\
 \hline
 
 \textbf{DinoGRL(our)}     &-                       &Video   &6    & \textbf{72.5} & \textbf{82.9} & \textbf{86.8} & \textbf{61.1}  & \textbf{76.1} & \textbf{85.3} & \textbf{87.9} & \textbf{62.3} \\ \toprule[0.8pt]
\end{tabular}
\end{table*}

\begin{table*}[!ht]\centering\small
\caption{Performance comparison with the state-of-the-art Re-ID methods on BUPTCampus. `R@1', `R@5' and `R@10' denote Rank-1,Rank-5 and Rank-10, respectively.}
\label{Tab2}
\begin{tabular}{m{2.3cm}<{\centering}m{2cm}<{\centering}m{1.5cm}<{\centering}m{1cm}<{\centering}m{0.7cm}<{\centering}m{0.7cm}<{\centering}m{0.7cm}<{\centering}m{0.7cm}<{\centering}m{0.7cm}<{\centering}m{0.7cm}<{\centering}m{0.7cm}<{\centering}m{0.7cm}<{\centering}}
\toprule[0.8pt]
\multirow{2}*{Methods} & \multirow{2}*{Reference} & \multirow{2}*{Type} & \multirow{2}*{Seq\_Len} & \multicolumn{4}{c}{\textit{Infrared to Visible}} & \multicolumn{4}{c}{\textit{Visible to Infrared}} \\ 
\cmidrule(lr){5-8} \cmidrule(lr){9-12} 
& & & & {R@1} & {R@5} & {R@10} & mAP & {R@1} & {R@5} & {R@10} & mAP \\ \toprule[0.8pt]

  LbA\cite{park2021learning}    &ICCV'21  &Image &10  & 32.1 & 54.9 & 65.1  & 32.9 & 39.1 & 58.7 & 66.5  & 37.1 \\

  CAJ\cite{ye2021channel}       &ICCV'21 &Image  &10  & 40.5 & 66.8 & 73.3  & 41.5 & 45.0 & 70.0 & 77.0  & 43.6  \\
  AGW\cite{ye2021deep}          &TPAMI'21 &Image &10  & 36.4 & 60.1 & 67.2  & 37.4 & 43.7 & 64.4 & 73.2  & 41.1 \\
  MMN\cite{zhang2021towards}    &CVPR'21 &Image  &10  & 40.9 & 67.2 & 74.4  & 41.7 & 43.7 & 65.2 & 73.5  & 42.8 \\
    DART\cite{yang2022learning}   &CVPR'22 &Image  &10  &52.4 &70.5 &77.8 &49.1  &53.3  &75.2  &81.7  &50.5\\
  DEEN\cite{zhang2023diverse}   &CVPR'23  &Image &10  &53.7 &74.8 &80.7 &50.4  &49.8  &71.6  &81.0  &48.6 \\

 \hline
 MITML\cite{VCM}    &CVPR'22  &Video &6  &49.1 &67.9 &75.4 &47.5 &50.2 &68.3 &75.7 &46.3   \\
 AuxNet\cite{BUPT}       &TIFS'24  &Video &10  &\underline{63.6} &\underline{79.9} &\underline{85.3} &\underline{61.1} &\underline{62.7} &\underline{81.5} &\underline{85.7} &\underline{60.2}    \\
 \hline

    \textbf{DinoGRL(our)}     &-               &Video  &6   &\textbf{61.8} &\textbf{81.6} &\textbf{84.8} &\textbf{60.1} &\textbf{65.2} &\textbf{82.6} &\textbf{86.5} &\textbf{61.1}   \\
   \textbf{DinoGRL(our)}     &-               &Video  &10   &\textbf{65.0} &\textbf{81.2} &\textbf{85.7} &\textbf{62.2}     &\textbf{70.3} &\textbf{86.9} &\textbf{89.8} &\textbf{64.1}   \\ \toprule[0.8pt]
\end{tabular}
\end{table*}

\textbf{Progressive Local-to-Global Enhancement. }
For each granularity $s$, the global appearance feature $\textbf{f}\hspace{0.1em}^i_{a,m}$ is progressively enhanced by sequentially interacting with all local gait stripe features $\{\textbf{f}\hspace{0.1em}^{i,j}_{g,m,s}\}_{j=1}^{n_s}$. This sequential enhancement consists of $n_s$ iterative steps.
Initially, the global appearance feature is projected into a latent embedding space to produce the initial query representation:
\begin{equation}
Q^0_s = \psi^q_s(\textbf{f}\hspace{0.1em}^i_{a,m}),
\end{equation}
where $\psi^q_s(\cdot)$ is a learnable 1D convolution layer for query embedding at granularity $s$.
At the $j$-th enhancement step ($j=1,\ldots,n_s$), the global feature $Q^{j-1}_s$ interacts with the local stripe $\textbf{f}\hspace{0.1em}^{i,j}_{g,m,s}$ through Local-to-Global Enhancement(LGE), producing an enhanced $Q^{j}_s$. specifically, LGE first projects the local stripe $\textbf{f}\hspace{0.1em}^{i,j}_{g,m,s}$ into key and value embeddings:
\begin{equation}
K^j_s = \psi^k_s(\textbf{f}\hspace{0.1em}^{i,j}_{g,m,s}), \quad V^j_s = \psi^v_s(\textbf{f}\hspace{0.1em}^{i,j}_{g,m,s}),
\end{equation}
where $\psi^k_s(\cdot)$ and $\psi^v_s(\cdot)$ are 1D convolution layers specific to granularity $s$.
The cross-modal relation between the global query and the local stripe is then modeled by computing the attention maps:
\begin{equation}
M_s^{j,+} = \text{ReLU}({Q^{j-1}_s}^\top K^j_s ), \quad M_s^{j,-} = \text{ReLU}(- {Q^{j-1}_s}^\top K^j_s),
\end{equation}
where $M_s^{j,+}$ captures the positively correlated identity-consistent components, and $M_s^{j,-}$ captures the negatively correlated modality-specific noise.
Using the attention maps, the enhancement vector contributed by the $j$-th stripe is computed as:
\begin{equation}
\Delta \textbf{f}^j_s = V^j_s \times M_s^{j,+} - V^j_s \times M_s^{j,-},
\end{equation}
which injects complementary gait cues into the global appearance representation.
The global feature $Q_s^{j-1}$ is then updated by incorporating the enhancement vector:
\begin{equation}
Q^j_s = Q^{j-1}_s + \Delta \textbf{f}^j_s,
\end{equation}
where $Q^j_s$ serves as the updated query for the next LGE step.
After completing all $n_s$ steps, the final enhanced global appearance feature at granularity $s$ is obtained as:
\begin{equation}
\overline{\textbf{f}}\hspace{0.1em}^i_{a,m,s} = Q^{n_s}_s.
\end{equation}

A similar sequential enhancement process is applied to the gait stream, where the global gait feature ${\textbf{f}\hspace{0.1em}_{g,m}^i}$ is progressively updated by interacting with local appearance stripes $\{\textbf{f}\hspace{0.1em}^{i,j}_{a,m,s}\}_{j=1}^{n_s}$. The final enhanced global gait feature at granularity $s$ is obtained as $\overline{\textbf{f}}\hspace{0.1em}^i_{g,m,s} = Q^{n_s}_s$.

\textbf{Multi-Granularity Identity Supervision.}  
For each enhanced global feature at granularity $s$, identity classification is independently performed for both streams. The overall identity loss is formulated as:
\begin{equation}
\mathcal{L}_{identity} = \frac{1}{N_s} \sum_{g=1}^{N_s} \left( \mathcal{L}_{app}^{g} + \mathcal{L}_{gait}^{g} \right),
\end{equation}
where $N_s=5$ denotes the total number of supervised representations, including the original global feature and the four enhanced global features obtained from different spatial granularities. Overall the proposed PBMGE module effectively bridges the appearance and gait streams through fine-grained bidirectional interactions and hierarchical aggregation, leading to more robust, modality-invariant, and detail-preserving pedestrian representations.

\subsection{Optimization}\label{sec3.4}

The training is performed in an end-to-end manner. The multi-granularity identity loss $\mathcal{L}_{identity}$ ensures identity discriminability for sequence-level pedestrian representations.
To preserve the semantic priors from DINOv2 while enabling adaptive optimization for ReID, a regularization loss $\mathcal{L}_{mask}$ is employed.
Meanwhile, the combination of $\mathcal{L}_{smo}$ and $\mathcal{L}_{div}$ encourages $\bm E_a$ and $\bm E_g$ to extract diverse and semantically rich features.
The overall training objective is:
\begin{equation}
\mathcal{L}_{total} = \mathcal{L}_{identity} + \lambda_1 \mathcal{L}_{mask} + \lambda_2 \mathcal{L}_{smo} + \lambda_3 \mathcal{L}_{div},
\end{equation}
where $\lambda_1$, $\lambda_2$, and $\lambda_3$ are balancing hyperparameters.

\section{EXPERIMENTS}
\subsection{Datasets and Experimental Settings}\label{sec4.1}

\hspace{1em} \textbf{Datasets.} We evaluate our method on two public VVI-ReID datasets: \textbf{HITSZ-VCM}\cite{VCM} and \textbf{BUPT}\cite{BUPT}. HITSZ-VCM contains 927 identities with 251,452 RGB and 211,807 IR images, organized into 11,785 visible and 10,078 infrared tracklets. BUPT includes 3,080 identities, 1,869,066 images, and 16,826 trajectories, averaging 111 images per trajectory.

\textbf{Evaluation metrics.} The standard Cumulative Matching Characteristics (CMC) curve and mean Average Precision (mAP) are adopted as the evaluation metrics.

\textbf{Implementation details.} All experiments are conducted on a single NVIDIA Quadro RTX 8000 GPU with PyTorch framework. We adopted ResNet50 pre-trained on ImageNet as the backbone, with input images resized to 256×128 pixels. A learning rate warmup strategy is used, starting at 0.1 and decayed to 0.01 and 0.001 at the 35th and 80th epochs, respectively. Training runs for 200 epochs, with hyperparameters $\lambda_{1}$, $\lambda_{2}$ and $\lambda_{3}$ set to 1, 0.02 and 5. Data augmentation includes Random Crop, Random Horizontal Flip, Channel Random Erasing and Channel AdapGray \cite{ye2021channel}. Each mini-batch samples 8 identities, with 4 VIS and 4 IR sequences per identity.

\subsection{Comparasion with State-of-the-Art Methods }\label{sec4.2}

In this section, we compare DinoGRL with existing state-of-the-art VVI-ReID methods on the public VVI-ReID datasets HITSZ-VCM and BUPT. As shown in Tab~\ref{Tab1}, DinoGRL consistently outperforms previous methods on the \textbf{HITSZ-VCM} dataset, demonstrating its superior effectiveness. Similarly, results on the \textbf{BUPT} dataset, presented in Tab~\ref{Tab2}, show notable improvements over existing approaches. It is worth noting that, since different methods adopt varying default sequence lengths, we conduct experiments with sequence lengths of 6 and 10 to ensure fair comparison on BUPT.

\subsection{Ablation Study}\label{sec4.3}
In this subsection, we conduct ablation studies on HITSZ-VCM to show the effectiveness of our proposed DinoGRL framework.

\textbf{Contributions of Proposed Components:} Tab. \ref{Tab3} reports the ablation study on the key modules of DinoGRL, including PBMGE and SASGL.

Integrating SASGL into the baseline improves performance by generating semantically enriched and task-adaptive gait representations via SASG and joint learning strategy.
Adding PBMGE further enhances performance by progressively enhancing global features via multi-granularity bidirectional enhancement between appearance and gait features.
When combined, the two modules yield the best results, demonstrating their strong synergy and effectiveness.

\textbf{Loss component in SASGL: }We conduct ablation experiments on the HITSZ-VCM dataset to evaluate the contributions of each loss component in SASGL. As shown in Tab. \ref{Tab4}, both the $\mathcal{L}_{smo}$ and the $\mathcal{L}_{div}$ individually bring performance gains when added to the base loss $\mathcal{L}_{b}$, and jointly optimizing them achieves the best performance, demonstrating their complementary effect.

\begin{table}[t]
\small
\centering
\caption{Ablation studies of DinoGRL. `B': Baseline.}
\label{Tab3}
\begin{tabular}{ccccccc}
\toprule[0.8pt]
\multicolumn{3}{c}{Component} & \multicolumn{2}{c}{\textit{IR to VIS}} & \multicolumn{2}{c}{\textit{VIS to IR}} \\ \cmidrule(lr){4-5} \cmidrule(lr){6-7} 
                                   B &  PBMGE  &  SASGL  & R@1& mAP& R@1& mAP\\ \toprule[0.8pt]
\textcolor[rgb]{0,0.5,0}{\ding{51}} &\textcolor{red}{\ding{55}} &\textcolor{red}{\ding{55}} & 63.5 & 46.9 & 65.7 & 48.1  \\
\textcolor[rgb]{0,0.5,0}{\ding{51}} &\textcolor[rgb]{0,0.5,0}{\ding{51}} &\textcolor{red}{\ding{55}} & 69.1  & 54.9  & 72.8  & 55.1  \\
\textcolor[rgb]{0,0.5,0}{\ding{51}} &\textcolor{red}{\ding{55}} &\textcolor[rgb]{0,0.5,0}{\ding{51}} & 70.1 & 59.0 & 73.1 & 60.1 \\
\textcolor[rgb]{0,0.5,0}{\ding{51}} &\textcolor[rgb]{0,0.5,0}{\ding{51}} &\textcolor[rgb]{0,0.5,0}{\ding{51}} & 72.5  &61.1 &76.1 &62.3 \\ \toprule[0.8pt]
\end{tabular}
\end{table}

\begin{table}[t]
\centering\small
\caption{Ablation studies on Loss Components in SASGL. $\mathcal{L}_{b}=\mathcal{L}_{reid}+\mathcal{L}_{mask}$ denotes the base loss, which is indispensable.}
\label{Tab4}
\begin{tabular}{ccccccc}
\toprule[0.8pt]
\multirow{2}{*}{$\mathcal{L}_{b}$} & \multirow{2}{*}{$\mathcal{L}_{smo}$} & \multirow{2}{*}{$\mathcal{L}_{div}$} & \multicolumn{2}{c}{\textit{IR to VIS}} & \multicolumn{2}{c}{\textit{VIS to IR}} \\ \cmidrule(lr){4-5} \cmidrule(lr){6-7}  
                                    &    &    & R@1& mAP& R@1& mAP\\ \toprule[0.8pt]
\textcolor[rgb]{0,0.5,0}{\ding{51}} &\textcolor{red}{\ding{55}} &\textcolor{red}{\ding{55}} & 71.8 & 60.1 & 74.6 & 61.2  \\
\textcolor[rgb]{0,0.5,0}{\ding{51}} &\textcolor[rgb]{0,0.5,0}{\ding{51}} &\textcolor{red}{\ding{55}} & 72.1  & 60.3  & 75.1  & 62.2  \\
\textcolor[rgb]{0,0.5,0}{\ding{51}} &\textcolor{red}{\ding{55}} &\textcolor[rgb]{0,0.5,0}{\ding{51}} & 72.0 & 60.4 & 75.3 & 61.5 \\
\textcolor[rgb]{0,0.5,0}{\ding{51}} &\textcolor[rgb]{0,0.5,0}{\ding{51}} &\textcolor[rgb]{0,0.5,0}{\ding{51}} & 72.5  &61.1 &76.1 &62.3 \\ \toprule[0.8pt]
\end{tabular}
\end{table}

\begin{table}[t]
\centering\small
\caption{Ablation studies of the Gait and Appearance Encoders $E_g$ and $E_a$ in SASGL. }
\label{Tab5}
\begin{tabular}{m{1cm}<{\centering}m{1cm}<{\centering}m{0.7cm}<{\centering}m{0.7cm}<{\centering}m{0.7cm}<{\centering}m{0.7cm}<{\centering}}
\toprule[0.8pt]
\multirow{2}{*}{$E_g$} & \multirow{2}{*}{$E_a$} & \multicolumn{2}{c}{\textit{IR to VIS}} & \multicolumn{2}{c}{\textit{VIS to IR}} \\ 
\cmidrule(lr){3-4} \cmidrule(lr){5-6} 
                                      &    & R@1& mAP& R@1& mAP\\ 
\toprule[0.8pt]
\textcolor[rgb]{0,0.5,0}{\ding{51}} &\textcolor{red}{\ding{55}} & 68.4  & 55.9  & 70.4  & 57.2  \\
\textcolor{red}{\ding{55}} &\textcolor[rgb]{0,0.5,0}{\ding{51}} & 67.4 & 53.5 & 69.7 & 55.1 \\
\textcolor[rgb]{0,0.5,0}{\ding{51}} &\textcolor[rgb]{0,0.5,0}{\ding{51}} & 72.5  &61.1 &76.1 &62.3 \\ 
\toprule[0.8pt]
\end{tabular}
\end{table}

\begin{table}[t]
\centering\small
\caption{Ablation studies on the necessity of the Joint Learning Strategy in SASGL. }
\label{Tab6}
\begin{tabular}{ccccc}
\toprule[0.8pt]
\multirow{2}{*}{Methods} & \multicolumn{2}{c}{\textit{IR to VIS}} & \multicolumn{2}{c}{\textit{VIS to IR}} \\ \cmidrule(lr){2-3} \cmidrule(lr){4-5}  
                         & R@1        & mAP        & R@1        & mAP        \\ \toprule[0.8pt]
w/o Joint Learning       & 72.1       & 60.5       & 75.0       & 61.6       \\
w/ Joint Learning        & 72.5       & 61.1       & 76.1       & 62.3       \\ \toprule[0.8pt]
\end{tabular}
\end{table}

\begin{table}[t]
\centering\small
\caption{Comparison of Upstream Models for SASGL. }
\label{Tab7}
\begin{tabular}{m{2cm}<{\centering}m{0.7cm}<{\centering}m{0.7cm}<{\centering}m{0.7cm}<{\centering}m{0.7cm}<{\centering}}
\toprule[0.8pt]
\multirow{2}{*}{Methods} & \multicolumn{2}{c}{\textit{IR to VIS}} & \multicolumn{2}{c}{\textit{VIS to IR}} \\ \cmidrule(lr){2-3} \cmidrule(lr){4-5}   
                         & R@1        & mAP        & R@1        & mAP        \\ \toprule[0.8pt]
SCHP\cite{li2020self}                     & 68.7       & 53.3       & 71.8       & 54.7       \\
Grapy-ML\cite{he2020grapy}                 & 69.1       & 54.9       & 72.8       & 55.1       \\ \hline
DINOv2               & 72.5       & 61.1       & 76.1       & 62.3       \\ \toprule[0.8pt]
\end{tabular}
\end{table}

\textbf{Effect of $\bm E_g$, $\bm E_a$ and joint learning strategy in SASGL: }To validate the necessity of key components within SASGL, we conduct ablation studies on the HITSZ-VCM dataset. As shown in Tab. \ref{Tab5}, using either $\bm E_g$ or $\bm E_a$ alone leads to significant performance drops, as each encoder captures only modality-specific patterns and fails to leverage their complementarity. Further, Tab. \ref{Tab6} shows that removing the joint learning strategy also degrades performance. 

\textbf{Comparison of Upstream Models:} We utilized DINOv2 as the upstream model in SASGL to provide general-purpose visual priors. To validate its effectiveness, we compared it with several alternative upstream models: (1) SCHP\cite{li2020self}, a widely used parsing network, and (2) Grapy-ML\cite{he2020grapy}, a multi-level representation learning model. As shown in Tab. \ref{Tab7}, DINOv2 achieves superior performance over the other upstream models, demonstrating its advantage in providing robust and generalizable features for SASGL. 

\textbf{Impact of Granularity Number in PBMGE:} To determine the optimal setting, we vary the granularity number from 2 to 16. As shown in Tab. \ref{Tab8}, performance consistently improves with more granularities, demonstrating that richer multi-granularity information enhances feature learning. The best results are achieved at 16 granularities, which we adopt as the default configuration.

\textbf{Impact of Attention Modules within PBMGE:} We introduce a dedicated attention mechanism, LGE, in PBMGE to enhance multi-granularity feature interaction. To validate its effectiveness, we compared it with several commonly used feature interaction designs. As shown in Tab. \ref{Tab9}, `replacing LGE with simple feature concatenation (“w/o attention”) or other existing designs yields limited or even degraded performance. In contrast, LGE achieves the best results, demonstrating its necessity within PBMGE.

\begin{table}[t]
\centering\small
\caption{Ablation studies of spatial Granularities in PBMGE.}
\label{Tab8}
\begin{tabular}{m{2cm}<{\centering}m{0.7cm}<{\centering}m{0.7cm}<{\centering}m{0.7cm}<{\centering}m{0.7cm}<{\centering}}
\toprule[0.8pt]
\multirow{2}{*}{\textit{Granularity}} & \multicolumn{2}{c}{\textit{IR to VIS}} & \multicolumn{2}{c}{\textit{VIS to IR}} \\ 
\cmidrule(lr){2-3} \cmidrule(lr){4-5}  
                             & R@1  & mAP  & R@1  & mAP  \\ 
\toprule[0.8pt]
2                            & 70.6 & 58.8 & 73.3 & 60.1 \\
4                            & 72.4 & 60.5 & 75.7 & 61.8 \\
8                            & 72.0 & 60.2 & 75.6 & 61.5 \\
16                           & 72.5 & 61.1 & 76.1 & 62.3 \\ 
\toprule[0.8pt]
\end{tabular}
\end{table}

\begin{table}[t]
\centering\small
\caption{Ablation Study of Attention in PBMGE.}
\label{Tab9}
\begin{tabular}{m{2cm}<{\centering}m{0.7cm}<{\centering}m{0.7cm}<{\centering}m{0.7cm}<{\centering}m{0.7cm}<{\centering}}
\toprule[0.8pt]
\multirow{2}{*}{Methods} & \multicolumn{2}{c}{\textit{IR to VIS}} & \multicolumn{2}{c}{\textit{VIS to IR}} \\ \cmidrule(lr){2-3} \cmidrule(lr){4-5}   
                                                          & R@1        & mAP        & R@1        & mAP        \\ \toprule[0.8pt]
w/o attention                                                  & 70.6       & 59.2       & 73.3       & 60.0       \\
Nonlocal\cite{wang2018non}                                & 69.3       & 57.3       & 73.7       & 58.8       \\
AttnFusion\cite{fan2024skeletongait}                      & 69.2       & 56.0       & 71.4       & 57.1       \\
IBAN\cite{IBAN}                                           & 70.3       & 57.7       & 74.0       & 58.5       \\
SCRL\cite{li2023shape}                                    & 69.2       & 55.9       & 71.7       & 56.9       \\
MS-G3D\cite{liu2020disentangling}                         & 63.5       & 49.1       & 66.7       & 49.6       \\ \hline
Ours                                                       & 72.5       & 61.1       & 76.1       & 62.3       \\ \toprule[0.8pt]
\end{tabular}
\end{table}

\begin{figure}[t!]
\centering
\includegraphics[width=7.8cm,keepaspectratio=true]{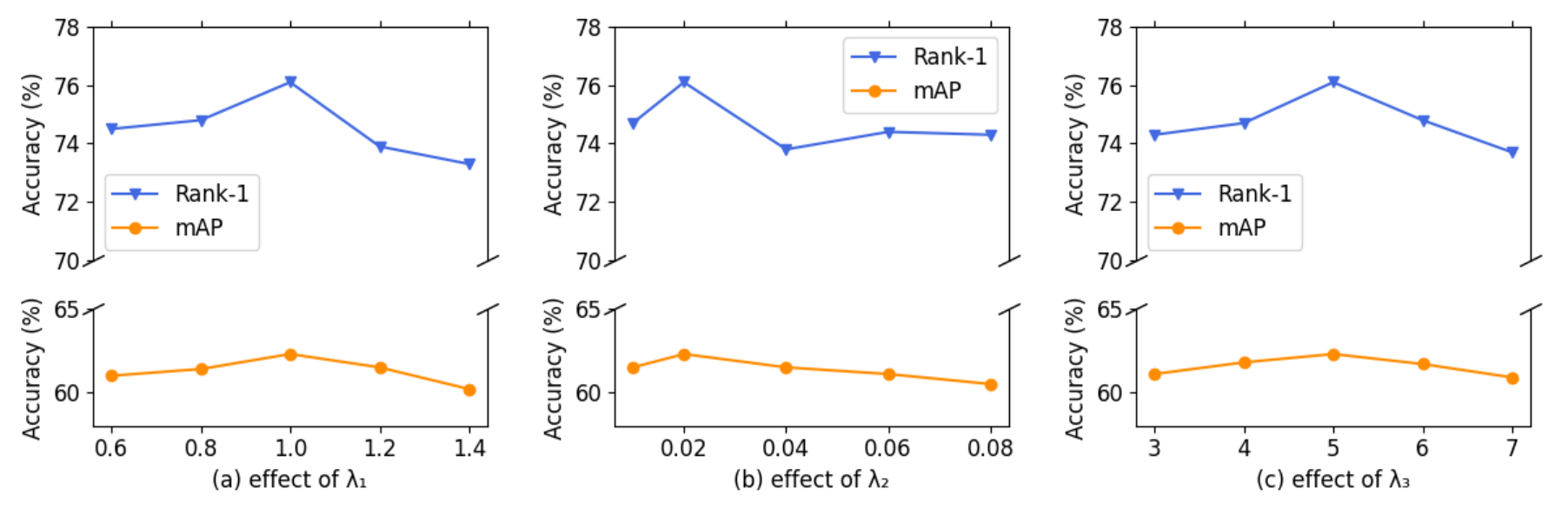}
\caption{Results of Rank-1 and mAP with different values of $\lambda_{1}$, $\lambda_{2}$ and $\lambda_{3}$ on HITSZ-VCM dataset. }
\label{param}
\end{figure}

\textbf{Impact of weight $ \lambda_1, \lambda_2, \lambda_3$ in the Objective Function:} We investigate the effects of the hyperparameters $ \lambda_1, \lambda_2, \lambda_3$ in the total loss function. As shown in Fig. \ref{param}, we vary $\lambda_1$ from 0.6 to 1.4, $\lambda_2$ from 0.01 to 0.08, $\lambda_3$ from 3 to 7. Specifically, $\lambda_1=1.0$, $\lambda_2=0.02$, $\lambda_3=5.0$ achieve the highest Rank-1 accuracy and mAP.

\section{VISUALIZATION}
To qualitatively assess the effectiveness of DinoGRL, we visualize the retrieval results and feature distance distributions.

\begin{figure}[t!]
\centering
\includegraphics[width=7.8cm,keepaspectratio=true]{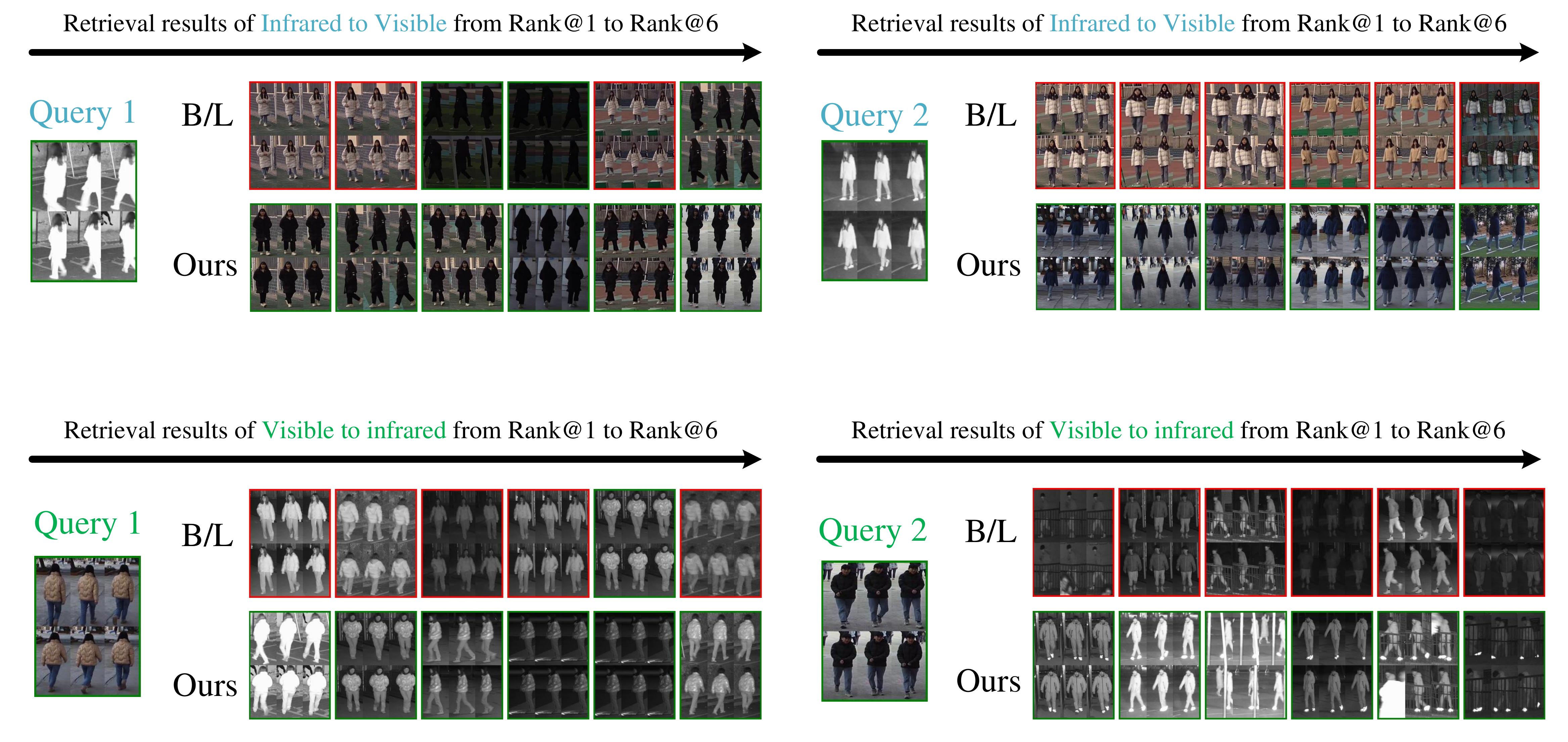}
\caption{Pedestrian search results (Top-6 results; B/L: baseline; green: correct match; red: incorrect match. }
\label{retrieval}
\end{figure}

\textbf{Retrieval Results Analysis.} Retrieval examples in Fig.~\ref{retrieval} show that the baseline, relying solely on appearance, suffers from cross-modal errors, such as misinterpreting white regions in IR images. In contrast, DinoGRL fully leverages complementary features, achieving more accurate retrieval even under challenging modality shifts.

\begin{figure}[t!]
\centering
\includegraphics[width=7.8cm,keepaspectratio=true]{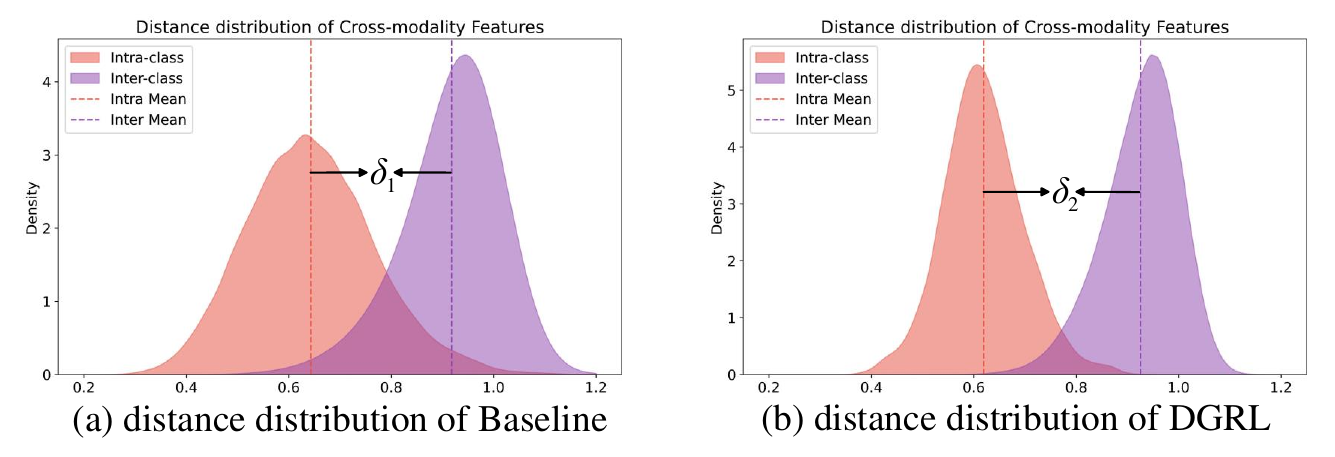}
\caption{Visualization of Feature Distance Distributions Between baseline and DGRL, where $\delta_2>\delta_1$.}
\label{dis}
\end{figure}

\textbf{Feature Distribution Analysis.} Fig.~\ref{dis} shows that DinoGRL achieves clearer intra- and inter-class separation compared to the baseline, demonstrating its superior discriminative capability.

\section{CONCLUSION}
This paper presents the DINOv2-Driven Gait Representation Learning (DinoGRL), a framework that leverages DINOv2’s general-purpose visual priors and the complementary strengths of appearance and gait to learn discriminative and modality-robust representations. We propose the Semantic-Aware Silhouette and Gait Learning (SASGL) model, which generates high-quality gait representations guided by DINOv2’s semantic priors and jointly optimizes them for task-adaptive ReID learning. Furthermore, the Progressive Bidirectional Multi-Granularity Enhancement (PBMGE) module refines features through multi-granularity interactions. Extensive experiments on HITSZ-VCM and BUPT datasets demonstrate that DinoGRL achieves state-of-the-art performance in VVI-ReID.

\begin{acks}
This research was supported by the National Natural Science Foundation of China (Nos. 62362045, 61966021, 62276120), the Basic Research Project of Yunnan Province (No. 202401AT070412), and the Yunnan Fundamental Research Projects (Nos. 202301AV070004, 202401AS070106). 
\end{acks}

\bibliographystyle{ACM-Reference-Format}
\balance
\bibliography{sample-base}

\end{document}